\RequirePackage[hyphens]{url}

\documentclass{article}

\usepackage{booktabs} 
\usepackage{microtype}
\usepackage{subcaption}
\usepackage{multicol}
\usepackage[dvipsnames]{xcolor}
\usepackage{hyperref}
\usepackage{amsthm}

\usepackage[accepted]{icml2019}
\usepackage{xcolor}
\usepackage{graphicx}
\usepackage{bm}
\usepackage{amsmath}
\usepackage{amssymb}
\usepackage{caption}
\usepackage{subcaption}
\usepackage{cleveref}
\usepackage{diagbox}
\usepackage{booktabs}
\usepackage{multirow,tabularx}
\usepackage{threeparttable}
\usepackage{xspace}
\usepackage{tabularx}
\usepackage[normalem]{ulem}
\usepackage{multirow}
\usepackage{tikz}
\usepackage{bbding}
\usepackage{pifont}
\usepackage{wasysym}
\usepackage{amssymb}
\PassOptionsToPackage{hyphens}{url}\usepackage{hyperref}

\crefname{chapter}{Chapter}{Chapters}
\crefname{section}{$\S$}{$\S$}
\crefname{subsection}{$\S$}{$\S$}
\crefname{subsubsection}{$\S$}{$\S$}
\crefname{figure}{Fig}{Figs}
\crefname{equation}{Eqn}{Eqns}
\crefname{table}{Table}{Tables}

\newcolumntype{Y}{>{\centering\arraybackslash}X}

\newif\ifdevelop\developfalse

\newif\ifcompact\compactfalse
\newif\ifnotcompact\notcompacttrue

\ifcompact
  \usepackage[skip=5pt]{caption}
  \usepackage[skip=3pt]{subcaption}
\else
  \usepackage{caption}
  \usepackage{subcaption}
\fi

\usepackage{setspace}
\ifcompact
  
\fi

\graphicspath{{figures/}}

\ifdevelop
  
\else
  
\fi

\ifdevelop
  \newcommand\devcomment[1]{\textcolor{red}{***{#1}***}}
\else
  \newcommand\devcomment[1]{}
\fi

\ifdevelop
  \newcommand\maybedeleted[1]{\textcolor{blue}{#1}}
\else
  \newcommand\maybedeleted[1]{}
\fi

\ifcompact
  \ifdevelop
    \newcommand\compactdel[1]{{\textcolor{red}{#1}}}
    
  \else
    \newcommand\compactdel[1]{}
    
  \fi
  \newcommand\compactvspace[1]{\vspace{#1}}
\else
  \newcommand\compactdel[1]{#1}
  
  \newcommand\compactvspace[1]{}
\fi

\newcommand{\disabled}[1]{}

\newcommand{\vv}[1]{\bm{\mathrm{{#1}}}}

\newcommand{\EEE}[2]{\operatorname{\mathbb{E}}_{{#1}}\left[{#2}\right]}

\newcommand{\KLDD}[2]{\operatorname{KL}\left[{#1}\,\big\|\,{#2}\right]}

\icmltitlerunning{Semi-Supervised Learning of Bearing Anomaly Detection via Deep Variational Autoencoders}

\begin{document}

\twocolumn[
\icmltitle{Semi-Supervised Learning of Bearing Anomaly Detection\\ via Deep Variational Autoencoders}



\icmlsetsymbol{equal}{*}
\begin{icmlauthorlist}
\icmlauthor{Shen Zhang}{gt}
\icmlauthor{Fei Ye}{path}
\icmlauthor{Bingnan Wang}{merl}
\icmlauthor{Thomas G. Habetler}{gt}
\end{icmlauthorlist}
\icmlaffiliation{gt}{School of Electrical and Computer Engineering, Georgia Institute of Technology, Atlanta GA, USA}
\icmlaffiliation{path}{California PATH, University of California, Berkeley, Berkeley CA, USA}
\icmlaffiliation{merl}{Mitsubishi Electric Research Laboratories, Cambridge MA, USA}

\icmlcorrespondingauthor{Shen Zhang}{\texttt{shenzhang@gatech.edu}}

\vskip 0.3in
]

%
\printAffiliationsAndNotice

\begin{abstract}
Most of the data-driven approaches applied to bearing fault diagnosis up to date are established in the supervised learning paradigm, which usually requires a large set of labeled data collected \emph{a priori}. In practical applications, however, obtaining accurate labels based on real-time bearing conditions can be far more challenging than simply collecting a huge amount of unlabeled data using various sensors. In this paper, we thus propose a semi-supervised learning approach for bearing anomaly detection using variational autoencoder (VAE) based deep generative models, which allows for effective utilization of dataset when only a small subset of data have labels. Finally, a series of experiments is performed using both the Case Western Reserve University (CWRU) bearing dataset and the University of Cincinnati's Center for Intelligent Maintenance Systems (IMS) dataset. The experimental results demonstrate that the proposed semi-supervised learning scheme greatly outperforms two mainstream semi-supervised learning approaches and a baseline supervised convolutional neural network approach, with the overall accuracy improvement ranging between 3\% to 30\% using different proportion of labeled samples.
\end{abstract}

%


\section{Introduction}
Electric machines are widely employed in a variety of industry applications and electrified transportation systems, and they consume approximately 60\% of all electric power produced. However, on certain occasions these machines may operate at unfavorable conditions, such as high ambient temperature, high moisture, and overload, which can eventually result in motor malfunctions that lead to high maintenance costs, severe financial losses, and various safety hazards. Among the many types of electric machine failures, it is revealed that the bearing fault is the most common fault type that accounts for 30\% to 40\% of all the machine failures, according to several surveys conducted by the IEEE Industry Application Society (IEEE-IAS) \cite{IAS} and the Japan Electrical Manufacturers' Association (JEMA) \cite{JEMA} 

In the last few years, many data-driven approaches have been applied to enhance the accuracy and reliability of bearing fault diagnosis \cite{review1, review2}. Specifically, many time-series signals used for monitoring the bearing condition have high-dimensional feature spaces, which makes it ideal to exploit deep learning algorithms to perform feature extraction and classification. While many models have presented satisfactory results after proper training sessions, most of them are established in the supervised learning paradigm, which requires a large set of labeled data collected \emph{a priori} under all possible fault conditions.

For bearing anomaly detection, however, it is often times much easier to collect a large amount of data than to accurately obtain their corresponding labels, and this is especially the case for faults that evolved naturally over time. For instance, even though an electromechanical system will be eventually turned off due to the consequences of a fault, it is not easy to determine precisely when the first trace of such a fault actually shows up and how long it lasts at the incipient stage. Additionally, it may also suffer from data ambiguity issues due to challenges associated with properly labeling these data, especially at a transition stage when they exhibit early signs of fault features but are far from obvious when compared to those features of a fully developed fault. Therefore, for accuracy concerns, only the very beginning and the very end of the collected data can be confidently assigned to their corresponding labels, leaving many of the intermediate data unlabeled, which apparently cannot be exploited using supervised learning. 

To train a bearing fault classifier with a good level of generalization using only a small subset of labeled data, one approach is to apply specific algorithms that can make the most out of the labeled data and computing resources available. For example, data augmentation techniques such as generative adversarial networks (GAN) \cite{GANOrigin}, have already applied in \cite{GAN1} and \cite{GAN2}, as well as some cross validation methods such as leave-one-out and Monte Carlo. Additionally, another promising route is to apply the semi-supervised learning paradigm to leverage the manually labeled data and the massive unlabeled data.

Semi-supervised learning specifically considers the problem of classification when only a small subset of data have corresponding labels, and as of now only a handful of semi-supervised learning paradigms have been applied to bearing anomaly detection. For example, a support vector data description (SVDD) model proposed in \cite{SVDD} utilizes cyclic spectral coherence domain indicators to build the feature space and fits a hyper-sphere, which later calculate the Euclidean distances in order to isolate data from healthy and faulty conditions. Moreover, graph-based methods are employed in \cite{graph1} and \cite{graph2} to construct a graph connecting similar samples in the dataset, so the class labels can propagate through the graph from labelled to unlabelled nodes. However, these approaches are sensitive to the graph structure and require eigen-analysis of the graph Laplacian, which limits the scale to which these methods can be applied. Instead of graph-based methods, \cite{S3AL} captures the structure of the data using $\alpha$-shape, which is mostly utilized for the surface estimation that reduces parameter tuning. 

Additionally, the recent semi-supervised deep ladder network \cite{ladder_origin} is implemented in \cite{ladder} to identify one-stage parallel shaft helical gear faults in induction machine systems. The ladder network is accomplished through the integration of the supervised and unsupervised learning strategies by modeling hierarchical latent variables. However, the unsupervised component of the ladder network may not be helpful in the semi-supervised setting if the raw data do not exhibit significant clustering on the 2D manifold, which is often not the case for bearing vibration signals. While GANs can be also implemented for semi-supervised learning, it has been reported in \cite{GAN_semi} that a good semi-supervised classifier and a good generator cannot be obtained at the same time. In other words, good semi-supervised learning actually requires a bad generator, and the well-known difficulties to train GANs further adds to the uncertainty of applying them to semi-supervised learning tasks. 

The motivation for the proposed research is both broad and specific, as we seek to approach the problem of bearing anomaly detection with solid theoretical explanation, and leverage properties of both the labeled and unlabeled data to make the classifier performance more accurate than that based on the labelled data alone. Therefore, we adopt the semi-supervised learning algorithm as the deep generative model, which is built upon solid Bayesian theory and scalable variational inference. Although some prior work employing VAE can be found in \cite{VAE_feature1} and \cite{VAE_feature2}, they are only utilizing the discriminative feature in the latent space for dimension reduction and use these features to train other external classifiers, while in this work we also make use of the generative function of VAE and train itself as a classifier through an integrated approach. 
\section{Background of Variational Autoencoders}
\begin{figure}
	\centering
	\ifcompact
	  \includegraphics[height=0.65in]{variational-auto-encoder-horizon}
	\else
	  \includegraphics[height=1.3in]{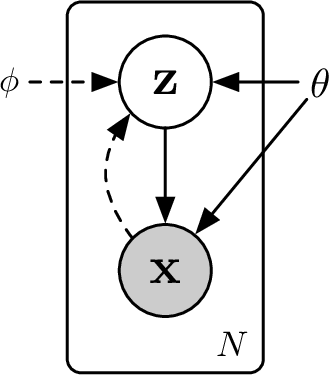}
	\fi
	\caption{Architecture of VAE.  The prior of $\vv{z}$ is regarded as part of the generative model (solid lines), thus the whole generative model is denoted as $p_{\theta}(\vv{x},\vv{z}) = p_{\theta}(\vv{x}|\vv{z})\,p_{\theta}(\vv{z})$. The approximated posterior (dashed lines) is denoted as $q_{\phi}(\vv{z}|\vv{x})$.}
	\label{fig:vae}
\end{figure}
\compactvspace{-.5em}
Variational inference techniques \cite{VAE} are often adopted in the process of training and prediction, which are efficient methods to estimate posteriors of the distributions derived by neural networks. The architecture of VAE is shown in \cref{fig:vae}, which is formulated in a probabilistic perspective that specifies a joint distribution $p_{\vv{\theta}}(\vv{x},\vv{z})$ over the observed variables $\vv{x}$ and latent variables $\vv{z}$, such that $p_{\vv{\theta}}(\vv{x},\vv{z})=p_{\vv{\theta}}(\vv{x}|\vv{z})p(\vv{z})$. Latent variables are drawn from a prior density $p(\vv{z})$ that is usually a multivariate unit Gaussian $\mathcal{N}(\vv{0},\vv{I})$, and these local latent variables $\vv{z}$ are related to their corresponding observations $\vv{x}$ through the likelihood $p_{\vv{\theta}}(\vv{x}|\vv{z})$, which can be viewed as a probabilistic decoder (generator) to decode $\vv{z}$ into a distribution over the observation $\vv{x}$. The exact form of decoder $p_{\theta}(\vv{x}|\vv{z})$ can be modeled as a neural network with parameters $\vv{\theta}$ consisting of the weights $\vv{W}$ and biases $\vv{b}$ of this network. 

Having specified the decoding process, it is also desirable to perform inference, or to compute the posterior $p_{\vv{\theta}}(\vv{z}|\vv{x})$ of latent variables $\vv{z}$ given the observed variables $\vv{x}$. Additionally, we also wish to optimize the model parameters $\vv{\theta}$ with respect to $p_{\vv{\theta}}(\vv{x})$ by marginalizing out the latent variables $\vv{z}$ in $p_{\vv{\theta}}(\vv{x},\vv{z})$. However, since the true posterior $p_{\vv{\theta}}(\vv{z}|\vv{x})$ is analytically intractable due to the Gaussian non-conjugate prior $p(\vv{z})$, then the variational inference technique is used to find an approximate posterior $q_{\vv{\phi}}(\vv{z}|\vv{x})$ with optimized variational parameters $\vv{\phi}$ that minimizes its Kullback-Leibler (KL) divergence to the true posterior \textcolor{blue}{(add references here)}. This approximated posterior $q_{\vv{\phi}}(\vv{z}|\vv{x})$ can be also viewed as an encoder that is usually assumed to be $\mathcal{N}(\vv{z}|\vv{\mu}_{\phi}(\vv{x}),\operatorname{diag}(\vv{\sigma}^2_{\phi}(\vv{x})))$, where values of $\vv{\mu}_{\phi}(\vv{x})$ and $\vv{\sigma}_{\phi}(\vv{x})$ are optimized by neural networks. 

Due to similar intractability issues such that computing the KL divergence explicitly would require the log marginal likelihood $\log p_{\vv{\theta}}(\vv{x})$, we'll maximize an alternative objective function known as the evidence lower bound (ELBO) of this log likelihood, which is defined as
\vspace{-0.05in}
\begin{equation}
    \begin{aligned} 
        \log p_{\theta}(\vv{\vv{x}}) &\geq \underbrace{\log p_{\theta}(\vv{x}) - \KLDD{q_{\phi}(\vv{z}|\vv{x})}{p_{\theta}(\vv{z}|\vv{x})}}_{\mathrm{ELBO}} \label{eqn:vae-elbo}
    \end{aligned}
\end{equation}

Specifically, the ELBO is a lower bound on the probability of data given a specific model, which can be further written as 
\begin{equation}
\begin{aligned} 
\operatorname{ELBO} &=\mathbb{E}_{q_{\phi}(\mathbf{z} | \mathbf{x})}\left[\log p_{\theta}(\mathbf{x} | \mathbf{z})+\log p_\theta(\mathbf{z})-\log q_{\phi}(\mathbf{z} | \mathbf{x})\right] \\ &=\mathbb{E}_{q_{\phi}(\mathbf{z} | \mathbf{x})}\left[\log p_{\theta}(\mathbf{x} | \mathbf{z})\right]-\mathrm{KL}\left[q_{\phi}(\mathbf{z} | \mathbf{x}) \| p_\theta(\mathbf{z})\right] 
\end{aligned}
\label{eqn:3}
\end{equation}

Therefore, maximizing it with respect to model parameters $\vv{\theta}$ approximately maximizes the log marginal likelihood. Additionally, maximizing it with respect to variational parameters $\vv{\phi}$ is equivalent to minimizing the KL divergence. The maximization of ELBO requires its gradients with respect to $\vv{\theta}$ and $\vv{\phi}$, which are also generally intractable. Currently, the dominant approach for circumventing this is by Monte Carlo (MC) integration \cite{geweke1989bayesian} to approximate the expectation of the gradients, as shown in \cref{eqn:monte-carlo-integration}, where $\vv{z}^{(l)}, l=1 \dots L$ are samples from $q_{\phi}(\vv{z}|\vv{x})$, and then perform stochastic gradient descent with the repeated MC gradient estimates.
\begin{equation}
	\EEE{q_{\phi}(\vv{z}|\vv{x})}{f(\vv{z})}
		\approx \frac{1}{L} \sum_{l=1}^L f(\vv{z}^{(l)})
	\label{eqn:monte-carlo-integration}
\end{equation}

\section{Deep Generative Model Architecture for Semi-supervised Learning}

This section presents two models for semi-supervised learning \cite{VAE_semi} that exploit the generative functionality of VAE to improve the classification performance when only a small subset of data have labels. The labeled data has the form of $(\vv{x_l}, y_l)$, and the unlabeled data is represented by $\vv{x_u}$.

By learning a close variational approximation of the posterior, the encoder part of the VAE is able to provide an embedding or feature representation of the input data $\vv{x}$ as a set of latent features $\vv{z}$. The probabilities of the approximated posterior $q_{\phi}(\vv{z}|\vv{x})$ are formed by nonlinear transformations that can be modeled as a deep neural network $f(\vv{z};\vv{x},\vv{\phi})$ with variational parameters $\vv{\phi}$. On the other hand, the generator part of VAE takes a set of latent variables $\vv{z}$ and produces new observations $\vv{x'}$ using a suitable likelihood function $p_{\vv{\theta}}(\vv{x}|\vv{z})$, which can be modeled as another deep neural network $g(\vv{x};\vv{z},\vv{\theta})$ with model parameters $\vv{\theta}$. 
\compactvspace{-.6em}
\subsection{Latent-feature discriminative model (M1)}
\begin{figure}
	\centering
	\includegraphics[width=\columnwidth]{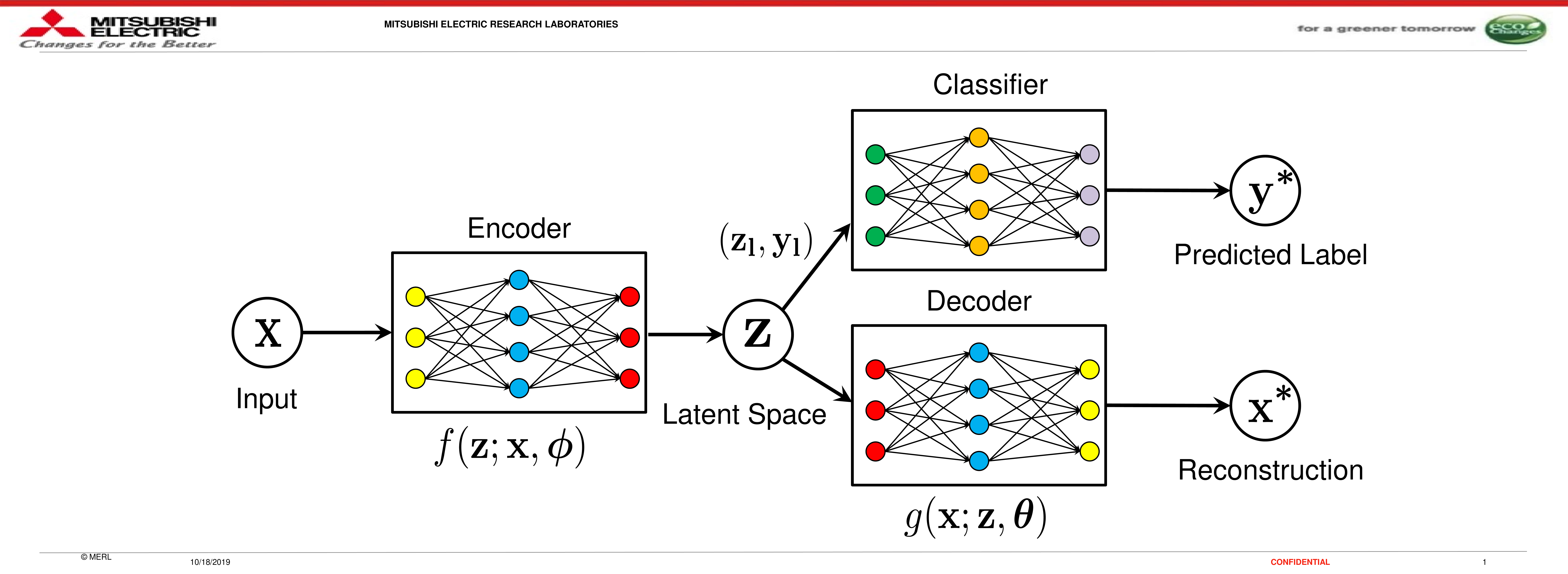}
	\caption{Illustration of the latent-feature discriminative model(M1).
	}\label{fig:M1_model}
\end{figure}
\compactvspace{-.5em}
A very straightforward implementation of VAE in semi-supervised learning, also referred to as the M1 model in \cite{VAE_semi}, is to train the model using both the labeled and unlabeled training data. When both the encoder and generator parameters are optimized, only the encoder part is utilized to provide a clustering of input data in the latent space. In most cases, the dimension of latent space variables $\vv{z}$ is much smaller than that of the input data $\vv{x}$, and these low-dimensional embeddings can be handled more easily in a supervised learning problem.

After training VAE, the classification task will be handed over to an external classifier. Only the labeled data $\vv{x_l}$ will be processed by the VAE encoder to determine their corresponding latent variables $\vv{z_l}$, which will be used as input features to train this classifier, such as a (transductive) SVM or multinomial regression, aided by their corresponding labels $y_l$, as shown in Fig. \ref{fig:M1_model}. Since the M1 model is trained using all data available in an unsupervised manner, it can generally promote a more accurate classification while only using a limited number of labeled data. 
\begin{figure*}
\begin{align*}
\operatorname{ELBO_U} &=  \mathbb{E}_{q_\phi(y, {\bf z}|{\bf x})}\bigg[
    \log p_{\theta}({\bf x}|y, {\bf z}) + \log p_{\theta}(y)
      + \log p_{\theta}({\bf z}) - \log q_\phi(y, {\bf z}|{\bf x})
\bigg] \\
&= \mathbb{E}_{q_\phi(y|{\bf x})}\bigg[
   \mathbb{E}_{q_\phi({\bf z}|{\bf x})}\big[
    \log p_{\theta}({\bf x}|y, {\bf z})
   \big]
    + \log p_{\theta}(y)
    - KL[q_{\phi}({\bf z}|{\bf x})||p_{\theta}({\bf z})]
    - \log q_{\phi}(y|{\bf x})
\bigg] \\       \tag{6}
&= \mathbb{E}_{q_\phi(y|{\bf x})}\big[ -\mathcal{L}({\bf x}, y))
    - \log q_{\phi}(y|{\bf x})
    \big] \\
&= \sum_y q_\phi(y|{\bf x})(-\mathcal{L}({\bf x}, y))
    + \mathcal{H}(q_\phi(y|{\bf x})) = -\mathcal{U}(x)
\label{eqn:6}
\end{align*}
\begin{align*} 
\operatorname{ELBO_L} = & \mathbb{E}_{q_{\phi}(\mathbf{z}| \mathbf{x}, y)}\left[\log p_{\theta}(\mathbf{x} | y, \mathbf{z})+\log p_{\theta}(y)+\log p_{\theta}(\mathbf{z})-\log q_{\phi}(\mathbf{z} | \mathbf{x}, y)\right] \\
= & \mathbb{E}_{q_{\phi}(\mathbf{z} | \mathbf{x})}\left[\log p_{\theta}(\mathbf{x} | y, \mathbf{z})+\log p_{\theta}(y)+\log p_{\theta}(\mathbf{z})-\log q_{\phi}(\mathbf{z} | \mathbf{x})\right] \\ \tag{7}
= &-\mathcal{L}(\mathbf{x}, y)
\end{align*}
\vspace{-0.1in}
\hrule
\end{figure*}

\compactvspace{-.5em}
\subsection{Generative semi-supervised model (M2)}

A major limitation of the M1 model lies its disjoint nature in firstly training the VAE and later the external classifier. Specifically, the initial VAE training stage of the M1 model is a complete unsupervised process that does not directly involve any of the scarce labels $y_l$, which is entirely separated from the later classifier training stage that actually takes $y_l$. Therefore, another semi-supervised M2 model was also proposed in \cite{VAE_semi} to address this problem. 
%

The M2 model can simultaneously deal with two cases: one where the labels are available, and one where they are not provided. Thus there are also two different approaches when constructing the approximate posterior $q$ as well as the variational objective.

\subsubsection{Variational Objective with Unlabelled Data}
\begin{figure}
	\centering
	\includegraphics[width=\columnwidth]{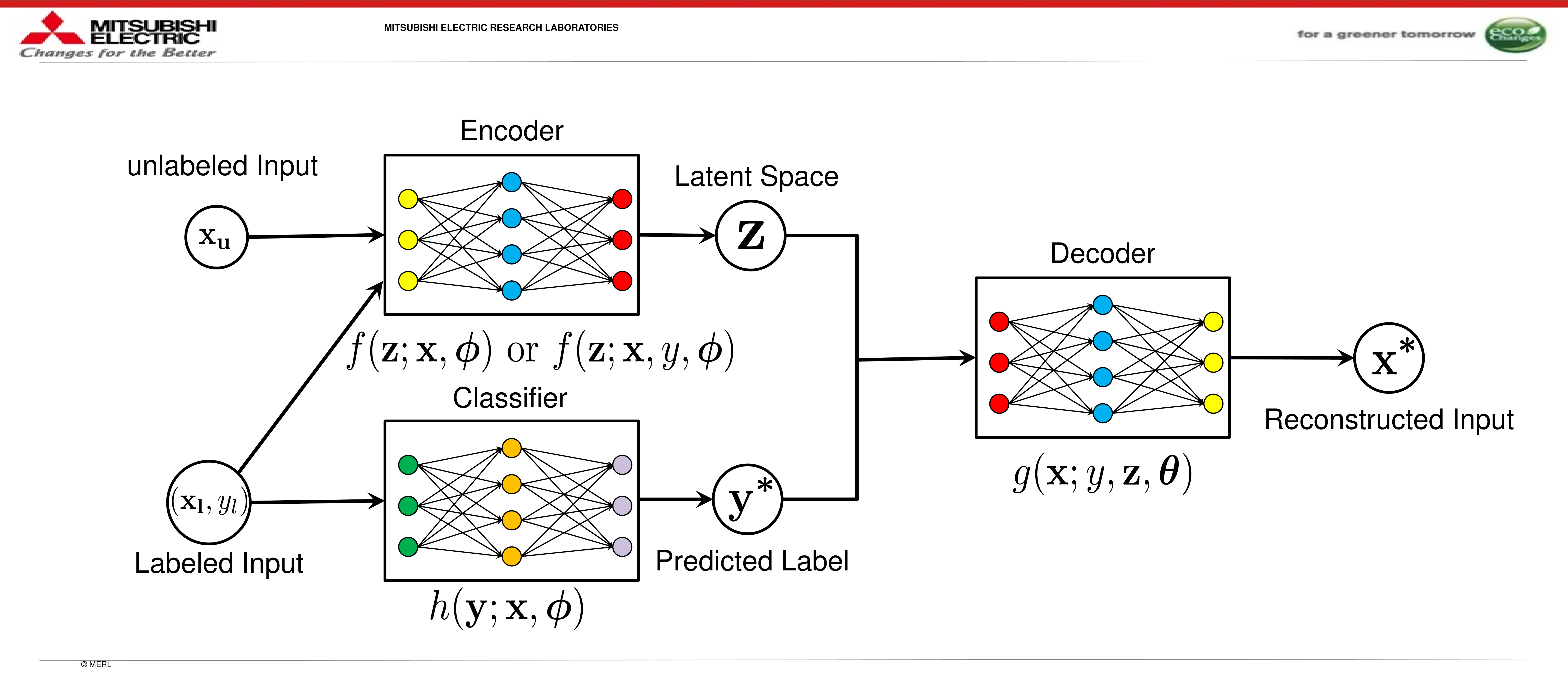}
	\caption{Illustration of the generative semi-supervised model (M2).
	}\label{fig:M2_model}
\end{figure}
For the case where labels are missing, two separate posteriors $q_{\vv{\phi}}(\vv{z}|\vv{x})$ and $q_{\vv{\phi}}({y}|\vv{x})$ will be involved in the VAE training phase, where $\vv{z}$ is still a vector latent variable similar to the M1 model, and ${y}$ is the unobserved label ${y_u}$. This newly defined posterior approximation $q_{\vv{\phi}}({y}|\vv{x})$ will be used to construct the best classifier possible as our inference model \cite{VAE_semi}. Specifically, the two approximated posteriors of class labels ${y}$ and latent variables $\vv{z}$ are defined as 
\begin{equation}
\begin{array}{l}{q_{\phi}(y|\mathbf{x})=\operatorname{Cat}\left(y | \pi_{\phi}(\mathbf{x})\right)} \\ {q_{\phi}(\mathbf{z} | \mathbf{x})=\mathcal{N}\left(\mathbf{z} | \mu_{\phi}(\mathbf{x}), \operatorname{diag}\left(\sigma_{\phi}^{2}(\mathbf{x})\right)\right)}\end{array}
\end{equation}
where $\operatorname{Cat}\left(y|\pi_{\phi}(\mathbf{x})\right)$ is the concatenated multinomial distribution, $\pi_{\phi}(\mathbf{x})$ is similar to $\mu_{\phi}(\mathbf{x})$ and $\sigma_{\phi}(\mathbf{x})$ in such a way that it is also defined by a neural network parameterized by $\phi$, which we need to learn from the training phase. By combining the above two posteriors, a joint approximated posterior can be defined as
\begin{equation}
q_{\phi}(y, \mathbf{z} | \mathbf{x})=q_{\phi}(\mathbf{z} | \mathbf{x}) q_{\phi}(y | \mathbf{x})
\end{equation}

Similar to Eqn. \ref{eqn:3}, the revised $\mathrm{ELBO_U}$ that determines the variational objective for the unlabeled data can be written as Eqn. 6, where $\mathcal{L}({\bf x}, y)$ is the original ELBO defined in Eqn. \ref{eqn:3}, 
\subsubsection{Variational Objective with Labelled Data}
Since the goal of semi-supervised learning is to train a classifier using a limited amount of labelled data combined with a vast majority of unlabeled data, it would be better to also incorporate the scarce labels to train the VAE classifier, which is unfortunately a featuer that cannot be accomplished using the M1 model. Similar to Eqn. \ref{eqn:3}, the revised $\mathrm{ELBO_L}$ that determines the variational objective for the labeled data can be written in the form of Eqn. 7.

\subsubsection{Combined Generative Semi-Supervised Model Objective}
As can be observed in Eqn. 6, the distribution $q_{\vv{\phi}}({y}|\vv{x})$, which is used to construct the discriminative classifier, is only contained in the variational objective of the unlabeled data. This is still an undesirable property since the labeled data will not be involved to learn this distribution and thus the variational parameter $\vv{\phi}$. Therefore, an extra loss term should be superimposed onto the combined model objective, such that both the labeled and labeled data would contribute to the training process. Thus the final objective of the generative semi-supervised model is:
\begin{equation}
\mathcal{J}^\alpha=\sum_{\mathbf{x} \sim \tilde{p}_{\mathbf{u}}} \mathcal{U}(\mathbf{x}) + \sum_{(\mathbf{x}, y) \sim \tilde{p}_{l}} \left[\mathcal{L}(\mathbf{x}, y)- \alpha \cdot\log q_{\phi}(y | \mathbf{x})\right] \tag{8}
\end{equation}
where the hyper-parameter $\alpha$ controls the relative weight between the generative and purely discriminative learning. A rules of thumb is to set $\alpha$ to be $\alpha = 0.1\cdot N$ in all experiments, with $N$ being the number of labeled data samples.

With this objective function, we can implement a sufficiently large number of $\vv{x}$ as a mini-batch to enhance the stability of training two neural networks that work as an encoder or a decoder. The objective function can be properly evaluated using either labelled or unlabelled data, and finally we run gradient descents to update the network parameters $\vv{\theta}$ and the variational parameters $\vv{\phi}$. The network of the M2 model is very close to a vanilla VAE with an additional posterior on $y$ and two additional terms on the objective function. An illustration of the M2 generative semi-supervised model is presented in Fig. \ref{fig:M2_model}.
\subsection{Model Implementations}
\subsubsection{VAE based M1 Model Implementations}
The VAE based M1 model consists of two deep neural networks $f(\vv{z};\vv{x},\vv{\phi})$ and $g(\vv{x};\vv{z},\vv{\theta})$ to represent the encoder $q_{\phi}(\vv{z}|\vv{x})$ and decoder $p_{\theta}(\vv{x}|\vv{z})$. The encoder structure has 2 convolutional layers and 1 fully connected layer with batch normalization, dropout and ReLU activation. The decoder consists of 1 fully connected layer followed by 3 transpose convolutional layers, among which the first 2 layers adopts ReLU activation and the final layer uses linear activation.

While training this VAE based M1 model, it is often times difficult to train a straight implementation of VAE that equally weighted the likelihood and the KL divergence, as shown in Eqn. 2, due to the ``KL-vanishing" problem as the KL loss can undesirably reduce to zero (although it is expected to maintain a small value). To overcome this, the implementation of M1 model uses the ``KL cost annealing" or ``$\beta$ VAE" \cite{KL_aneal1}, which has a modified ELBO function as 

\begin{equation}
\operatorname{ELBO}=\underbrace{\mathbb{E}_{q_{\phi}(z | x)}\left[\log p_{\theta}(x | z)\right]}_{\text {Reconstruction}}-\beta \cdot \underbrace{\operatorname{KL}\left(q_{\phi}(z | x) \| p(z)\right)}_{\text {KL Regularization}}
\end{equation}
in which the introduced weight factor of the KL divergence term $\beta$ will gradually increase from 0 to 1 over the course of the training epoch. When $\beta<1$, the latent variable $\vv{z}$ is trained to focus more on capturing useful information for reconstruction of $\vv{x}$. When the full VAE objective is considered ($\beta=1$), $\vv{z}$ learned in the earlier epoch earlier can be viewed as VAE initialization; such latent features are much more informative than the random start in constant schedule and thus are ready for the decoder to use \cite{KL_aneal2}

After training the M1 model that is able to balance its reconstruction and generation features, latent variables $\vv{z}$ in the latent space will be used as discriminative features that can be passed along to an external classifier. In this paper, the SVM classifier is adopted. While it is flexible to use any classifier of personal preference, the merit of the M1 model is to take advantage of the feature extraction functionality of VAE to reduce the dimension of the input data (1,024), to a much lower feature dimension, which is selected to be 128 in this study. 
\subsubsection{VAE based M2 Model Implementations}
The VAE based M2 model adopts the same structure for $q_{\phi}(\vv{z}|\vv{x})$ as in the M1 model, and the decoder $p_{\theta}(\vv{x}|y, \vv{z})$ also has the same settings as M1 $p_{\theta}(\vv{x}|\vv{z})$. The classifier $q_{\phi}(y|\vv{x})$ has 2 convolutional layers and 2 max pooling layers with dropout and ReLU activation, followed by a final Softmax layer.  

Two separate neural networks, one for the labeled data and one for the unlabeled data, are employed with identical network structure but different input/output specifications and loss functions \cite{VAE_blog}. For example, for the labeled data, both $\vv{x_l}$ and $y$ are treated as input to minimize the labeled ${(\mathbf{x}, y) \sim \tilde{p}_{l}}$ part in Eqn. 8, and the output would be the reconstructed $\vv{x_l'}$ and $y'$, while for the unlabeled data, $\vv{x_u}$ serves as input to generate the reconstructed $\vv{x_u'}$.

Other hyper-parameters for the M2 model are also chosen empirically. We use 200 as the batch size for training, and run for 10 epochs. The dimension of the latent variable $\vv{z}$ is 128. We use the RMSprop optimizer with an initial learning rate of $10^{-4}$.
\section{Experimental Evaluation}
\label{sec:experiments}
In this section, we seek to validate the effectiveness of the variational autoencoder based deep generative model for semi-supervised learning in the context of bearing anomaly detection. The formulation of the diagnosis framework will be presented in detail, and the classifier performance will be compared against other popular semi-supervised learning methods, such as PCA and the vanilla autoencoder (AE), as well as CNN that represents a benchmark supervised learning paradigm.
\subsection{Datasets}
In this experiment, both the Case Western Reserve University (CWRU) bearing dataset and the University of Cincinnati's Center for Intelligent Maintenance Systems (IMS) datasets are adopted, which contain data from manually initiated and naturally evolved bearing defects, respectively.
\subsubsection{CWRU Dataset}
The CWRU dataset contains mechanical vibration signals collected from the drive-end bearing in a 2-hp induction motor dyno setup. Single point defects are manually created to the bearing inner raceway, the outer raceway, and the rolling element using electro-discharge machining. Different levels of fault severity are modeled using different defect diameters such as 7 mils, 14 mils, 21 mils, 28 mils, and 40 mils. Vibration data are collected with motor loads from 0 to 3 hp, and motor speeds from 1,720 to 1,797 rpm using two accelerometers installed at both the drive end and fan end of the motor housing at a sampling frequency of either 12 kHz or 48 kHz. The generated dataset is recorded and also made publicly available on the CWRU bearing data center website \cite{CWRU}.
\begin{table}[]
    \caption{Selected class labels from the CWRU bearing dataset}
    \resizebox{\linewidth}{!}{
\begin{tabular}{lcccccc}
\toprule
\multirow{2}{*}{Class label} & \multicolumn{3}{c}{Fault location} & \multicolumn{3}{c}{Fault diameter (mils)} \\ \cmidrule{2-7} 
                             & Ball     & IR     & OR    & 0.007        & 0.014        & 0.021       \\ \midrule
1                            & \checkmark         & --       & --      & \checkmark             & --             & --            \\
2                            & \checkmark         & --       & --      & --             & \checkmark             & --            \\
3                            & \checkmark         & --       & --      & --             & --              & \checkmark            \\
\midrule
4                            & --         & \checkmark       & --      & \checkmark             & --             & --            \\
5                            & --         & \checkmark       & --      & --             & \checkmark             & --            \\
6                            & --         & \checkmark       & --      & --            & --              & \checkmark            \\
\midrule
7                            & --         & --       & \checkmark      & \checkmark             & --             & --            \\
8                            & --         & --       & \checkmark      & --             & \checkmark             & --            \\
9                            & --         & --       & \checkmark      & --             & --             & \checkmark            \\
\midrule
10                           & \multicolumn{6}{c}{Normal}            \\
\bottomrule
\end{tabular}}
\label{tab:CWRU}
\end{table}

Since the purpose of the bearing diagnostic framework is to accurately reveal both the location and severity of a bearing defect, vibration data collected for the same fault type but at different speeds and loading conditions would be considered as having the same label. Based on this criterion, 10 classes are identified based on the location and size of the bearing defect, and a detailed list of all 10 scenarios with their defect size and locations are demonstrated in Table \ref{tab:CWRU}. 
\compactvspace{-.5em}
\begin{figure}
	\centering
	\includegraphics[width=\columnwidth]{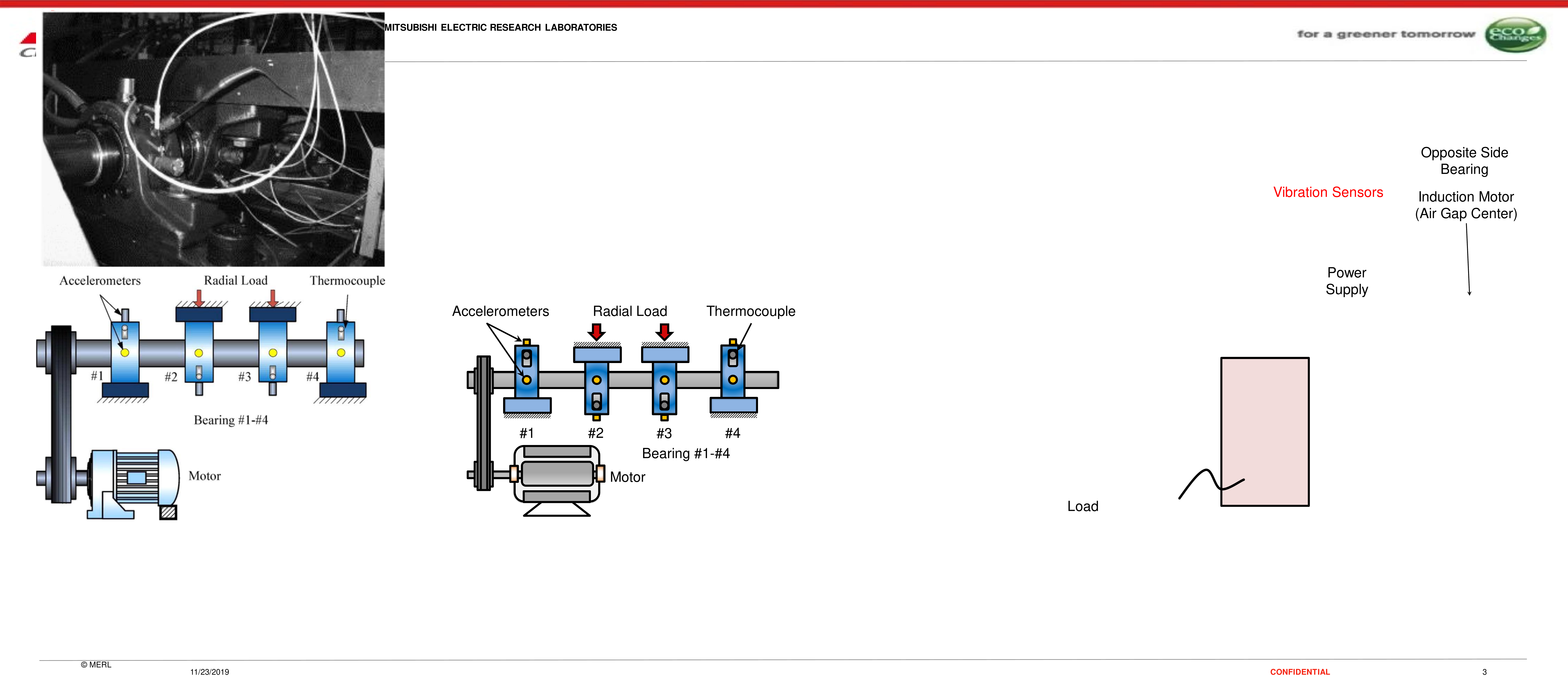}
	\caption{Experimental test rig for collecting the IMS dataset.}
	\label{fig:IMS}
\end{figure}
\begin{table*}[]
\caption{Bearing anomalies and degradation starting point of the IMS dataset}
\resizebox{\linewidth}{!}{
\begin{tabular}{lcccc}
\toprule
Bearing                    & Subset 1 Bearing 3 & Subset 1 Bearing 4 & Subset 2 Bearing 1 & Subset 3 Bearing 3 \\
\midrule
Fault type                 & Inner race         & Rolling element    & Outer race         & Outer race         \\
\midrule
Endurance duration         & 34 days 12 h               & 34 days 12 h               & 6 days 20 h                  & 45 days 9 h     \\
\midrule
Number of files         & 2,156               & 2,156               & 984                  & 4,448    \\
\midrule
Degradation starting point AEC$^*$ & 2,027               & 1,641               & 547                & 2367    \\
\midrule
Degradation starting point MAS-Kurtosis$^\dagger$ & 1,910               & 1,650               & 710                & N/A    \\
\midrule
Degradation starting point HMM-DPCA$^\ddagger$ & 2,120               & 1,760               & 539                & N/A    \\
\bottomrule
\vspace{-0.08in}
\end{tabular}}
\begin{footnotesize}
~~$^*$AEC: auto-encoder-correlation-based (AEC) prognostic algorithm.\\
~~$^\dagger$MAS-Kurtosis: moving average spectral kurtosis.\\
~~$^\ddagger$HMM-DPCA: hidden Markov model with dynamic PCA.
\end{footnotesize}
\label{tab:IMS}
\end{table*}
\subsubsection{IMS Dataset}
In most cases, bearing damage is artificially initiated in order to accelerate the degradation process. Therefore, the IMS dataset is also applied in this study to verify the effectiveness of VAE based generative models for detecting naturally evolved bearing defects. The IMS bearing dataset has been collected on an endurance test rig illustrated in Fig. \ref{fig:IMS}. Specifically, there are 4 double row bearings mounted on the same shaft, which is coupled to the motor shaft via a rubber belt. Each of the 4 bearings under test is applied with a radial load of 6,000 lbs, and since they are not used to support the motor or any active rotating movement, the endurance test can continue even these bearings have shown implications of failure. As a matter of fact, the stopping criterion for the test is when the accumulation of debris on a magnetic plug exceeded a certain level \cite{IMS}, which is a major difference when compared to a realistic scenario where the bearing is supporting the motor or transmission, and the test needs to be stopped quickly after spotting any anomaly conditions. 

The IMS data contains 3 subsets of data when the motor runs at a constant speed of 2,000 rpm, and there are two PCB 253B33 High sensitivity Quart ICP accelerometers installed on each bearing (x and y positions) for the first subset, while the other two subsets only have one accelerometer on each bearing. A 1-second acquisition has been made every ten minutes and stored in a separate data file that includes 20,480 sample points, with an an exception for the first subset for which the first 92 files have been acquired every five minutes. As briefly discussed earlier, since the endurance test may continue after bearing degradation, a summary of the bearing anomaly condition and the approximated degradation starting point is listed in Table \ref{tab:IMS}, where the starting points are estimated using 3 different algorithms adapted from \cite{IMS_start} .
\subsection{Data Preprocessing}
\subsubsection{Data Segmentation}
The diagnostic procedure begins with segmentation, during which the collected vibration signals are first divided into a number of segments with an equal length. 

For the CWRU dataset, the length of the vibration signal for each type of drive end motor bearing fault shown in Table \ref{tab:CWRU} is roughly 120,000 at three different speeds, namely 1,730 rpm, 1,750 rpm, and 1,772 rpm. Data collected at these speeds would constitute the complete fault data, which is later segmented according to a fixed window size of 1,024 (spans 85.3 ms in the dataset that was collected at 12 kHz) and a sliding ratio of 0.2. Ultimately the number of segments for the training data and the test data are 12,900 and 900, respectively. Although the percentage of test data might seem low (around 7\%) at the first glance, only a maximum of around 2,000 labeled training samples will be used in this semi-supervised learning setting, making the ratio of test data to the labeled training data above 40\%.

Similarly, the IMS dataset is also segmented using a fixed window size of 1,024 with data collected at 20.48 kHz. Due to a high level of noise present in the ``Subset 3 Bearing 3'' condition as reported in \cite{IMS_start}, only the first 3 fault conditions shown in Table \ref{tab:IMS} are selected for algorithm validation. In addition, in order to simulate a realistic scenario where the equipment would be stopped right after spotting a bearing anomaly condition, 210 files are selected for each fault type, with 15 of them chosen after their degradation starting points using the auto-encoder-correlation-based (AEC) prognostic algorithm. For example, for the ``Subset 1 Bearing 3'' condition, data files 1,832 to 2,042 would be selected, while the degradation starting point for this fault scenario is 2,027. On the other hand, the healthy data is selected as the first 110 files also from ``Subset 1 Bearing 3''. 

The last 10 files will be treated as the test set while the remaining 200 files form the entire training set for semi-supervised learning algorithms. As each files contains 20,480 data points that can be divided into 20 data segments, there are  altogether 4,000 data segments for each class, or 16,000 for all 4 classes (healthy, outer race, inner race, rolling element). The size of the test set is 800 data segments for all 4 classes. Similar to the way of constructing the training set and the test set, although the percentage of test data might seem low (around 5\%) at the first glance, this ratio is more significant when compared against the amount of labeled data, since only a limited number of the 160,000 training data segments will be labeled.

In order to simulate the challenges associated with data labeling for real world applications, the training set will be labeled starting from the last one of the 200 files, for which we should have the highest level of confidence regarding the accuracy of its label. Then more test cases will be created by labeling more preceding files but with a decreased level of confidence. The purpose is to validate if the inaccurate labeling would affect the accuracy of supervised learning algorithms, and if these data are left unlabeled, can they still promote the classifier accuracy using semi-supervised learning algorithms.
\compactvspace{-.5em}
\begin{figure*}
	\centering
	\includegraphics[width=\linewidth]{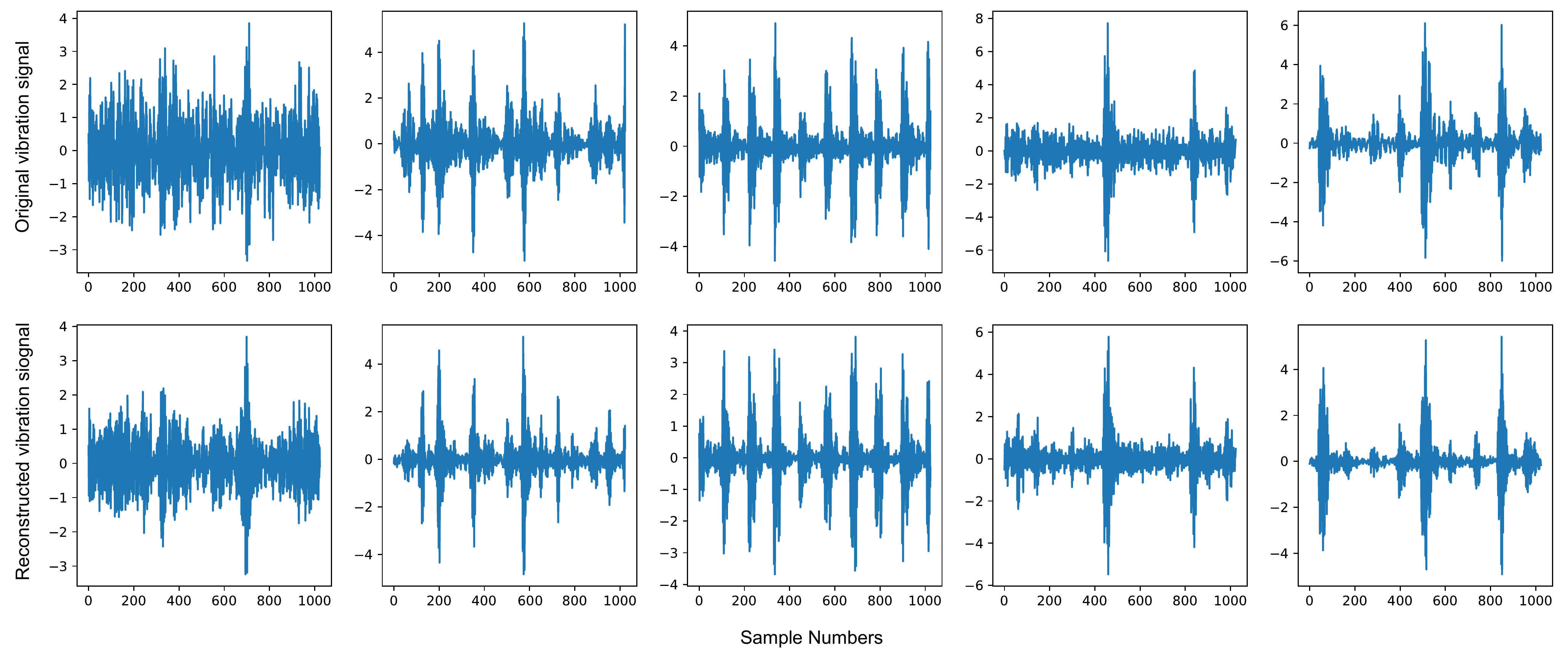}
	\caption{Comparison of the original and the reconstructed bearing vibration signals after training the VAE M1 model (top row: the original signal; bottom row: the reconstructed signal).}
	\label{fig:reconstruction}
\end{figure*}
\subsubsection{Data Shuffling and Standardization}
After the initial loading and segmentation of the data, the data segments are still sorted according to their class (fault types). Therefore, data shuffling needs to be performed to make sure that the training and test sets are representative of the overall distribution of the dataset, thus reducing the variance and making sure that the trained VAE models would remain general and come with less overfitting.

The conventional z-normalization technique is applied to both the training set and the test set to make sure the bearing vibration data in each segment have zero-mean and unit-variance. 
\subsection{Experimental Results}
Having successfully trained the VAE based model, the reconstructed bearing vibration signal should closely resemble the original signal, as shown in Fig. \ref{fig:reconstruction}. Although a perfect reconstruction (reconstruction error = 0) might cause a loss of generality and degrade the generative feature of VAE, a reasonably close reconstruction with a small error can indicate that the VAE has achieved a balance between reconstruction and generation, which is crucial for exploiting the generative feature of the algorithm.

Besides implementing the variational autoencoder based deep generative M1 and M2 model to perform bearing fault diagnostics with semi-supervised learning, a comparative study is also conducted in comparing with other popular semi-supervised learning methods, such as PCA and the traditional autoencoder, as well as mainstream supervised learning paradigms such as Convolutional Neural network (CNN). 
\begin{table*}[]
\centering
    \caption{Experimental results of semi-supervised classification on CWRU bearing dataset with limited labels}
\begin{tabular}{lccccc}
\toprule
$N$    & PCA+SVM              & Autoencoder      & CNN              & VAE M1           & VAE M2           \\
\midrule
10   & 17.10 ($\pm$ 1.25) & 27.72 ($\pm$ 1.28) & 21.33 ($\pm$ 2.52) & 23.04 ($\pm$ 1.32) & 23.71 ($\pm$ 1.67) \\
50   & 26.82 ($\pm$ 1.98) & 33.40 ($\pm$ 1.24) & 30.74 ($\pm$ 2.88) & 32.93 ($\pm$ 2.01) & 40.57 ($\pm$ 2.91) \\
100  & 32.79 ($\pm$ 1.99) & 37.96 ($\pm$ 0.65) & 34.50 ($\pm$ 2.16) & 36.91 ($\pm$ 1.37) & 60.04 ($\pm$ 3.57) \\
300  & 46.88 ($\pm$ 2.10) & 44.06 ($\pm$ 2.61) & 60.28 ($\pm$ 3.42) & 47.03 ($\pm$ 1.22) & 87.63 ($\pm$ 2.80) \\
516  & 50.56 ($\pm$ 1.69) & 50.88 ($\pm$ 2.03) & 75.41 ($\pm$ 2.74) & 57.06 ($\pm$ 1.76) & 94.16 ($\pm$ 1.66) \\
860  & 64.48 ($\pm$ 1.88) & 58.89 ($\pm$ 1.81) & 87.39 ($\pm$ 0.93) & 67.19 ($\pm$ 1.70) & 96.77 ($\pm$ 0.38) \\
1075 & 65.32 ($\pm$ 2.05) & 62.83 ($\pm$ 1.61) & 91.07 ($\pm$ 1.46) & 71.97 ($\pm$ 1.40) & 97.86 ($\pm$ 0.51) \\
2150 & 78.91 ($\pm$ 1.35) & 77.09 ($\pm$ 0.98) & 97.19 ($\pm$ 0.99) & 86.59 ($\pm$ 1.43) & 98.06 ($\pm$ 0.88) \\
\bottomrule
\end{tabular}
\label{tab:result}
\end{table*}
\subsubsection{CWRU Dataset}
For the CWRU dataset, a total of 10 rounds of semi-supervised experiments are performed on randomly shuffled training and test sets. Only a fraction of labels of the 129,000 training samples will be visible to different algorithms to construct a semi-supervised setting for bearing anomaly detection, where 8 case studies are carried out with 10, 50, 100, 245, 516, 860, 1,075, and 2,150 labels, representing roughly 0.08\%, 0.39\%, 0.78\%, 1.9\%, 4\%, 6.67\%, 8.33\%, and 16.67\% of the training data length, respectively.

The average accuracy and standard deviation for different algorithms after performing 10 rounds of experiments are presented in Tab. \ref{tab:result}, in which the latent-feature discriminative model (M1) performs better than other benchmark unsupervised models such as PCA and autoencoder, demonstrating the effectiveness of the latent space in providing robust features that allow for easier classification. It is worthwhile to notice that initially the VAE based M1 model compares favorably against CNN til the number of labeled samples $N=860$, after which it becomes slightly inferior, which is in contradiction to their implementation results on the MNIST dataset provided in \cite{VAE_semi}. One interpretation of this deviation might be due to the 1-D time-series CWRU dataset that has more explicit feature representations to be directly captured by supervised and fine-tuned structures such as CNN, as thus it may only require around 2,000 labeled training data segments to achieve a satisfactory result.

However, by combining features learnt in the M1 model with a classification mechanism directly in the same model, as in the conditional generative model (M2), we are able to get much better result than simply using a SVM classifier as in the M1 model. Specifically, it only needs 4\% of the training data to have labels to reach a fault classification accuracy of around 95\%, while the best value attainable from the rest of the algorithms is only 57.06\%, demonstrating a nearly 40\% increase in accuracy with very consistent performance, as its standard deviation is as low as 0.88\%.
\begin{table*}[]
\centering
    \caption{Experimental results of semi-supervised classification on IMS bearing dataset with limited labels}
\begin{tabular}{lccccc}
\toprule
$N$    & PCA+SVM              & Autoencoder      & CNN              & VAE M1           & VAE M2           \\
\midrule
10   & 56.74 ($\pm$ 0.44) & 54.10 ($\pm$ 1.58) & 53.92 ($\pm$ 2.77) & 32.93 ($\pm$ 3.81) & 52.79 ($\pm$ 1.98) \\
40   & 61.60 ($\pm$ 0.63) & 64.54 ($\pm$ 2.07) & 59.08 ($\pm$ 2.68) & 72.01 ($\pm$ 1.91) & 66.27 ($\pm$ 8.31) \\
100  & 67.22 ($\pm$ 1.15) & 67.07 ($\pm$ 1.49) & 62.93 ($\pm$ 4.10) & 76.61 ($\pm$ 1.48) & 71.15 ($\pm$ 6.24) \\
200  & 70.31 ($\pm$ 0.49) & 73.42 ($\pm$ 0.94) & 68.64 ($\pm$ 5.40) & 78.74 ($\pm$ 1.25) & 76.54 ($\pm$ 3.58) \\
400  & 75.38 ($\pm$ 0.90) & 78.42 ($\pm$ 1.17) & 74.20 ($\pm$ 3.17) & 81.66 ($\pm$ 1.02) & 82.78 ($\pm$ 2.21) \\
800  & 77.85 ($\pm$ 0.62) & 84.81 ($\pm$ 0.78) & 78.73 ($\pm$ 2.98) & 85.03 ($\pm$ 1.15) & 88.45 ($\pm$ 1.71) \\
1000 & 78.19 ($\pm$ 0.59) & 85.83 ($\pm$ 0.85) & 81.29 ($\pm$ 4.18) & 86.61 ($\pm$ 1.27) & 89.66 ($\pm$ 1.54) \\
2000 & 78.50 ($\pm$ 0.30) & 86.61 ($\pm$ 0.77) & 86.62 ($\pm$ 4.11) & 87.20 ($\pm$ 1.18) & 90.87 ($\pm$ 1.97) \\
4000 & 78.96 ($\pm$ 0.72) & 83.72 ($\pm$ 0.89) & 87.74 ($\pm$ 0.54) & 85.14 ($\pm$ 0.96) & 92.01 ($\pm$ 0.92) \\
8000 & 79.06 ($\pm$ 0.65) & 84.00 ($\pm$ 1.21) & 81.56 ($\pm$ 2.79) & 85.36 ($\pm$ 1.17) & 88.11 ($\pm$ 3.47) \\
\bottomrule
\end{tabular}
\label{tab:ims_result}
\end{table*}
\subsubsection{IMS Dataset}
Similarly, 10 rounds of semi-supervised experiments are performed on data retrieved from the IMS dataset, and 10 case studies are carried out by labeling the last 10, 40, 100, 200, 400, 800, 1,000, 2,000, 4,000, and 8,000 data segments of the training set, representing roughly 0.06\%, 0.25\%, 0.63\%, 1.25\%, 2.5\%, 5\%, 6.25\%, 12.5\%, 25\%, and 50\% of the training data, respectively. 

The average accuracy and standard deviation for the same 5 algorithms after performing 10 rounds of experiments are presented in Tab. \ref{tab:ims_result}. The latent-feature discriminative model (M1) performs better than PCA but has an almost equivalent performance as an autoencoder, which indicates that the discriminative latent space of VAE has no obvious advantage over the encoded feature space of vanilla autoencoders. However, the M1 model also outperforms the supervised learning algorithm CNN with a limited number of labeled data segments from $N=40$ to $N=1,000$ by approximately 5\% to 15\% with much smaller variance, demonstrating the benefit of incorporating the vast majority of unlabeled data in the training process.

Similar to the comparison result obtained using the CWRU dataset, the classifier performance of the deep generative VAE M2 model outperforms other 4 algorithms, showcasing the strength of the integrated training process of the VAE model and its built-in classifier. An important observation can be made that when the number of labeled data segments increases from $N=4,000$ to $N=8,000$,  the accuracy of CNN decreases by more than 6\%, while VAE M2 also suffers a 4\% loss. This can be largely attributed to the fact that the healthy data have been mistakenly labeled as faulty ones, although this decrease in accuracy was expected to happen while transitioning from $N=1,000$ to $N=2,000$, since there are only 1,000 faulty data segments before the anticipated bearing degradation starting point based on AEC as shown in Table \ref{tab:IMS}. This inconsistency might be caused by the accuracy of these estimated degradation starting points, since an early indication of the ``Subset 1 Bearing 3'' fault using the MAS-Kurtosis method is projected to be 2,320 data segments (9,280 for all 4 classes) earlier than the currently employed AEC results. Therefore, the ground-truth degradation point may lie somewhere in the middle between these two results. 

However, the experimental results obtained in the current setting are still able to support the earlier claims that taking advantage of the unlabeled data can effectively promote the classifier performance using the proposed VAE based semi-supervised learning methods, especially the deep generative M2 model. Additionally, it is also shown that inaccurate labeling can impair the accuracy of supervised learning algorithms in realistic applications, if the labels are assigned without a good level of confidence.

\compactvspace{-.3em}
\section{Conclusion}

In this paper, we applied two semi-supervised models based on the generative feature of VAE for bearing anomaly detection, which outperform state-of-the-art unsupervised and supervised learning algorithms. For the CWRU dataset, the improvement of the VAE M2 model with 4\% of labeled training data obtained a roughly 40\% accuracy enhancement. Considering the fact that the bearing defects are manually initiated in this dataset and are thus inconsistent with real-world scenarios that involve naturally evolved defects, we've also adopted the IMS dataset to validate the proposed semi-supervised VAE models. The results have shown that inaccurate labeling can degrade the classifier performance of the mainstream supervised learning algorithms, while incorporating semi-supervised models and keeping many of the miss-labeled data \emph{unlabeled} can be an effectively method improve the classifier performance.
%

%

\bibliographystyle{ACM-Reference-Format}
\bibliography{ref.bib}
\nocite{*}

\end{document}